\pdfoutput=1
\documentclass[10pt, a4paper]{article}

\usepackage[final]{lrec-coling2024} %

\usepackage{booktabs}
\usepackage{microtype}

\newenvironment{citemize}{\begin{list}{\tiny$\bullet$}{\topsep=.1\smallskipamount\itemsep=0pt\parsep=1pt\labelwidth=.5em}}{\end{list}}

\usepackage{xspace}
\def\evalatin{EvaLatin 2024 Dependency Parsing shared task\xspace}
\def\latinpipe{LatinPipe\xspace}

\title{ÚFAL LatinPipe at EvaLatin 2024: Morphosyntactic Analysis of Latin}

\name{Milan Straka, Jana Straková, Federica Gamba}

\address{Institute of Formal and Applied Linguistics, Faculty of Mathematics and Physics\\
Charles University, Czech Republic\\
\texttt{\{straka,strakova,gamba\}@ufal.mff.cuni.cz}}

\abstract{
We present \latinpipe, the winning submission to the \evalatin. Our system consists of a fine-tuned concatenation of base and large pre-trained LMs, with a dot-product attention head for parsing and softmax classification heads for morphology to jointly learn both dependency parsing and morphological analysis. It is trained by sampling from seven publicly available Latin corpora, utilizing additional harmonization of annotations to achieve a more unified annotation style. Before fine-tuning, we train the system for a few initial epochs with frozen weights. We also add additional local relative contextualization by stacking the BiLSTM layers on top of the Transformer(s). Finally, we ensemble output probability distributions from seven randomly instantiated networks for the final submission. The code is available at {\small\url{https://github.com/ufal/evalatin2024-latinpipe}}.
 \\ \newline \Keywords{dependency parsing, part of speech tagging, EvaLatin, Latin, LatinPipe} }

\usepackage[absolute]{textpos}
\begin{document}
\begin{textblock}{16}(0,0.1)\centerline{This paper was published at \textbf{LT4HALA 2024} -- please cite the published version {\small\url{https://aclanthology.org/2024.lt4hala-1.24/}}.}\end{textblock}

\maketitleabstract

\section{Introduction}
\label{sec:introduction}

In this paper, we describe our entry to the \evalatin \cite{sprugnoli-etal-2024-overview}. Our system is called \latinpipe to resemble its predecessors, UDPipe \cite{straka-strakova-2017-udpipe} and UDPipe 2 \cite{straka-2018-udpipe}. We submitted two variants, called \textit{ÚFAL LatinPipe~1} and \textit{ÚFAL LatinPipe~2}, placing 1st and 2nd in the shared task evaluation, respectively.

Our system is an evolution of UDPipe 2 \cite{straka-2018-udpipe}. \latinpipe is a graph-based dependency parser which uses a deep neural network for scoring the graph edges. Unlike UDPipe 2,  the neural network architecture of \latinpipe is a fine-tuned pre-trained language model, with a dot-product attention head for dependency parsing and softmax classification heads for morphological analysis to learn both these tasks jointly.

We provide an extensive evaluation of the approaches used in \latinpipe: a comparison of monolingual and multilingual pre-trained language models and their concatenations; initial pretraining on the frozen Transformer weights; adding two BiLSTM layers on top of the Transformers; and using the gold UPOS from the shared task data on the network input. A considerable focus is directed at multi-treebank training, as well as the harmonization of annotation styles among the seven publicly available Latin treebanks.

\section{Related Work}
\label{sec:related_work}
The \evalatin \cite{sprugnoli-etal-2024-overview} builds upon the two previous editions of EvaLatin, which focused respectively on lemmatization and POS tagging \cite{sprugnoli-etal-2020-overview} and lemmatization, POS tagging, and features identification \cite{sprugnoli-etal-2022-overview}. UDPipe 2 won the EvaLatin 2020 shared task \cite{straka-2020-evalatin}; previously, it participated  in the 2018 CoNLL Shared Tasks on Multilingual Parsing from Raw Text to Universal Dependencies \cite{zeman-etal-2018-conll}, which encompassed also Latin, and placed among the winning systems \cite{straka-2018-udpipe}.

\paragraph{Latin Dependency Parsing}
\looseness1
In recent years, \citet{nehrdich-hellwig-2022-accurate} developed a graph-based dependency parser specifically for Latin. Their approach modifies the architecture of the biaffine parser proposed by \citet{dozat2017deep} by incorporating a character-based convolutional neural network (CharCNN), and exploits Latin BERT embeddings \cite{bamman2020latin}.

\looseness1
\citet{fantoli-de-lhoneux-2022-linguistic} trained a POS tagging and parsing model using the deep biaffine parser \cite{dozat2017deep} implementation of MaChAmp \citep{van-der-goot-etal-2021-massive} and exploiting treebank embeddings in the encoder. 

\looseness1
\citet{karamolegkou-stymne-2021-investigation} explored Latin parsing in a low-resource scenario and found ancient Greek to be most effective as transfer language, likely due to its syntactic similarity with Latin.

\section{Data}
\label{sec:methods-data}

\paragraph{Latin Treebanks}

We train \latinpipe on the training portions of the following seven publicly available Latin corpora:

\begin{citemize}
    \item ITTB of UD 2.13 \citep{passarotti2019project};
    \item LLCT of UD 2.13 \citep{llct};
    \item PROIEL in either of these two versions: UD 2.13 \citep{proiel}, and a UD-style harmonized version (\citealp{gamba-zeman-2023-latin,gamba-zeman-2023-universalising});\footnote{Available for download at {\url{https://github.com/fjambe/Latin-variability/tree/main/morpho_harmonization/morpho-harmonized-treebanks}}.}
    \item UDante of UD 2.13 \citep{cecchini2020udante};
    \item Perseus of UD 2.13 \citep{perseus};
    \item UD-style annotated text of \textit{De Latinae Linguae Reparatione} by Marcus Antonius Sabellicus \citelanguageresource{gamba-cecchini-2024-de-latinae};
    \item \textit{Archimedes Latinus} UD-style treebank \citep{fantoli-de-lhoneux-2022-linguistic}, based on the Latin translation of the Greek mathematical work \textit{The Spirals} of Archimedes;\footnote{Available at \url{https://github.com/mfantoli/ArchimedesLatinus}.}
\end{citemize}

\begin{table}[]
    \centering
    \begin{tabular}{lr}
    \toprule
      Corpus & Training tokens \\
      \midrule
      ITTB   & 391K \\ %
      LLCT   & 194K \\ %
      PROIEL & 178K \\ %
      UDante & 31K \\ %
      Perseus & 18K \\ %
      Sab & 11K \\ %
      Arch & 1K \\ %
      \midrule
      UD 2.13 & 812K \\ %
      UD 2.13+Sab+Arch & 824K \\ %
      \bottomrule
    \end{tabular}
    \caption{Training data sizes in tokens.}
    \label{tab:training_data}
\end{table}

\noindent where UD 2.13 stands for the Universal Dependencies project \citep{nivre-etal-2020-universal}, version 2.13 \citelanguageresource{ud2.13}. We denote the former five corpora distributed by UD 2.13 as \textit{UD 2.13} and all seven corpora including additionally \textit{Arch} and \textit{Sab} as \textit{UD 2.13+Arch+Sab} in our experiments. The treebank training data sizes are presented in Table~\ref{tab:training_data}.

For the shared task, we train in multi-treebank setting, in which the examples from the abovementioned corpora are sampled into training batches proportionally to the square root of the number of their sentences, similarly to \citet{van-der-goot-etal-2021-massive}.

\paragraph{Harmonization of Annotation Styles}
\label{sec:methods-harmonization}

We noticed that the PROIEL treebank stands out most in terms of annotation style from the rest of the other treebanks, so much so that the differences in annotation style result in varying performance. We therefore experimented with the following three settings:

\begin{citemize}
    \item training with a harmonized version of PROIEL by \citet{gamba-zeman-2023-latin,gamba-zeman-2023-universalising}, submitted as \textit{ÚFAL LatinPipe~1};
    \item training without PROIEL altogether, submitted as \textit{ÚFAL LatinPipe~2};
    \item training with the original PROIEL annotation by \citet{proiel}, not submitted due to the two-runs-per-team limit.
\end{citemize}

The harmonized version of PROIEL resulted from the harmonization carried out by \citet{gamba-zeman-2023-latin,gamba-zeman-2023-universalising}, who observed persisting differences in the annotation scheme of the five Latin treebanks, annotated by different teams and in different stages of the development of UD guidelines. Divergences were observed at all annotation levels, from word segmentation to lemmatization, POS tags, morphology, and syntactic relations. The implemented harmonization process led to substantial improvements in parsing performances, confirming the need for a truly standardized annotation style.
Notably, among the five treebanks, in the case of PROIEL a lower degree of accordance with the UD guidelines was observed. For instance, in compound numerals like \textit{viginti quattuor} `twenty-four’ the second number is attached to the first one through a \texttt{fixed} relation; in the harmonized version, such dependencies are reannotated as \texttt{flat}. Moreover, PROIEL makes use of the \texttt{dep} relation, intended for cases where a more precise deprel cannot be assigned. Through POS tags and morphology, in the harmonized version \texttt{dep} is replaced with a more appropriate one.

\section{Methods}
\label{sec:methods}

\latinpipe is a graph-based dependency parser. First, a deep learning neural network is used to score the graph edge values, and then a global optimization Chu-Liu/Edmonds' algorithm \cite{chu-liu-1965,edmonds-1967-optimum} for finding the minimum spanning tree problem is run on the graph.

For scoring the graph edge values, \latinpipe pursues a deep learning approach and consists of a fine-tuned pre-trained LM (or a concatenation of them) with a dot-product parsing attention head. In addition, morphology softmax classification heads are also used, so \latinpipe jointly learns both dependency parsing and morphological analysis.

\begin{figure}[t]
    \centering
    \includegraphics[width=\hsize]{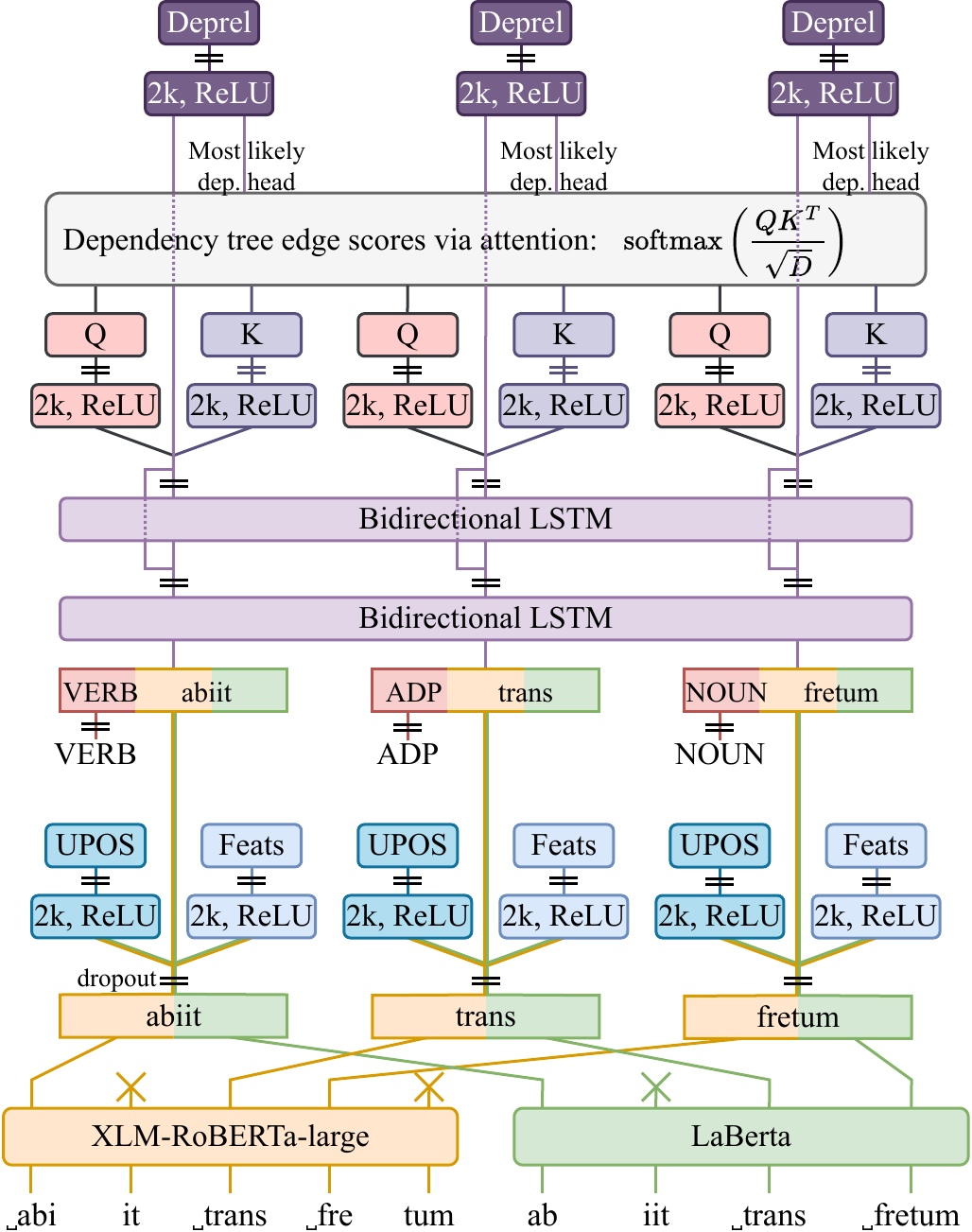}
    \caption{\latinpipe architecture overview.}
    \label{fig:architecture}
\end{figure}

The general overview of the architecture is given in Figure~\ref{fig:architecture} and the details are outlined in the following paragraphs.

\paragraph{Pre-trained LMs}

Our baselines are either fine-tuned LaBerta or PhilBerta, the Latin monolingual RoBERTa base language models by \citet{riemenschneider-frank-2023-exploring}; or the fine-tuned XLM-RoBERTa large (\citet{conneau-etal-2020-unsupervised}; 355M parameters), which was pretrained on 390M Latin tokens among other languages.
Apart from using the single fine-tuned PLMs, we also experimented with a concatenation of the contextualized embeddings yielded by multiple fine-tuned encoders: \textit{LaBerta+PhilBerta} and \textit{XLM-R large+LaBerta+PhilBerta}.

\paragraph{Frozen Pretraining}

Before fine-tuning the PLMs' weights, we can optionally freeze the pre-trained Transformer weights, and optimize solely the remaining weights of the architecture for a few initial epochs, namely the heads and the stacked BiLSTM layers. The objective of frozen pretraining is to facilitate the commencement of the fine-tuning optimization from a favorable starting point.

\paragraph{Adding LSTMs}

We incorporate two bidirectional LSTM layers \cite{hochreiter-schmidhuber-1997-lstm,gers-1999-learning} on top of the Transformer(s) to enhance the modeling of relative short-distance relationships between the tokens and to contextualize the embedded UPOS tags.

\paragraph{Gold UPOS on Input}

We leverage the gold morphological analysis provided in the shared task data as an additional input to the neural network. The trainable word embeddings of UPOS are concatenated with the contextualized embeddings yielded from the fine-tuned PLM(s), and together, the concatenation of embeddings is processed by the LSTM layers.

\paragraph{Ensembling}

For the final submission, we ensemble output probability distributions from seven randomly instantiated networks by averaging the probabilities in the corresponding dimensions.

\paragraph{Handling punctuation}

The shared task test data do not contain punctuation. This causes concern in settings when training without PROIEL, which is the only representative of a treebank without punctuation. Training solely on data containing punctuation is expected to lead to inferior performance on test data without it. Therefore in this particular setting, we artificially add punctuation to the test data by appending periods at sentence ends, and after the model prediction, we remove the dummy punctuation again.\footnote{Obviously, the other option would be to remove the punctuation from the training data and retrain the models, an expensive and unavailable option due to the restricted time span of the shared task testing period.}

\paragraph{Architecture Details}

In the LatinPipe architecture (Figure~\ref{fig:architecture}), every
classification layer and computation of queries and keys is preceded by
a hidden layer of size 2\,048 with ReLU activation. The dimensionality of the
queries and keys is 512, and the LSTM dimensionality is 256. When predicting
dependency relations, we also concatenate the LSTM-generated representation of
the most likely dependency head according to the predicted scores (which is not
necessarily the one chosen by the Chu-Liu/Edmonds' algorithm).

\paragraph{Training Details}

The model is trained with the Adam optimizer~\citep{kingma-ba-2015-adam} for 30 epochs, each comprising 1\,000 batches with a batch size of 32. The learning rate is first linearly increased from 0 to 2e-5 in the first two epochs and then decays to 0 according to the cosine schedule. Optionally, we perform 10-epoch pretraining with frozen Transformer weights utilizing a constant learning rate of 1e-3. On a single A100 GPU with 40GB, the training takes 9 hours. The exact training configuration, including the exact command used to train the models, is available in the released source code.

\section{Results}
\label{sec:results}

\begin{table*}
    \centering
    \catcode`! = 13\def!{\bfseries}
    \catcode`@ = 13\def@{\itshape}
    \def\0{\hphantom{0}}
    \setlength{\tabcolsep}{9.4pt}
    \renewcommand{\arraystretch}{0.945}
    \begin{tabular}{lcccccc}
      \toprule
      \textbf{Experiment} & Avg & ITTB & LLCT & \llap{P}ROIE\rlap{L} & \llap{U}Dant\rlap{e} & \llap{P}erseu\rlap{s} \\
      \midrule

  \multicolumn{7}{l}{\sc A) PLMs Evaluation}\vspace{0.1em}\\
    ~~LaBerta &  83.20 &  90.91 &  94.54 &  86.75 &  66.71 &  77.08 \\
    ~~PhilBerta &  82.87 &  91.09 &  94.19 &  86.13 &  66.42 &  76.51 \\
    ~~LaBerta+PhilBerta &  83.99 &  91.31 &  94.74 &  87.29 &  68.18 & !78.42 \\
    ~~XLM-R large &  84.19 &  91.60 &  95.33 &  87.18 &  71.17 &  75.67 \\
    ~~XLM-R large+LaBerta+PhilBerta & !84.67 & !91.78 & !95.35 & !87.57 & !71.95 &  76.70 \\
  \midrule
  \multicolumn{7}{l}{\sc B) Incremental Architecture Improvements w.r.t. the Previous Row}\vspace{0.1em}\\
    ~~+ Frozen training for 10 epochs &  86.09 &  92.29 &  95.34 &  88.64 &  74.20 &  79.98 \\
    ~~+ Two bidirectional LSTM layers &  86.33 &  92.81 &  94.70 &  89.05 &  74.78 &  80.32 \\
    ~~+ Gold UPOS on parser input & !86.97 & !93.18 & !95.64 & !89.78 & !74.99 & !81.28 \\
  \midrule
  \multicolumn{7}{l}{\sc C) Multi-treebank Training w.r.t. the Previous Row}\vspace{0.1em}\\
    ~~Single-treebank training &  86.97 & !93.18 & !95.64 & !89.78 &  74.99 &  81.28 \\
    ~~UD 2.13 training &  88.05 &  92.25 &  95.60 &  88.74 &  79.84 & !83.84 \\
    ~~UD 2.13+Sab+Arch training & !88.09 &  92.18 &  95.44 &  88.43 & !80.56 &  83.81 \\
  \midrule
  \multicolumn{7}{l}{\sc D) Ensembles of the Models in the Previous Section}\vspace{0.1em}\\
    ~~Single-treebank training, 7 models &  87.31 & !93.38 &  95.78 & !90.23 &  75.51 &  81.66 \\
    ~~UD 2.13 training, 7 models &  88.51 &  92.65 & !95.89 &  89.10 &  80.91 &  84.02 \\
    ~~UD 2.13+Sab+Arch training, 7 models & !88.63 &  92.45 &  95.78 &  89.23 & !81.47 & !84.22 \\
    \midrule
    \multicolumn{7}{l}{\sc E) Previous work}\vspace{0.1em}\\
    \multicolumn{2}{l}{~~@UDPipe 2 \cite{straka-2018-udpipe}, UD 2.12} &@89.35 &@94.39 &@79.55 & @68.65 &@71.91 \\
    \multicolumn{2}{l}{~~@MaChAmp \cite{van-der-goot-etal-2021-massive}, UD 2.8} &@92.45 &@95.41 &@86.97 & @74.01 &@74.67 \\
    \multicolumn{2}{l}{~~@\citet{nehrdich-hellwig-2022-accurate}, UD 2.8-2.9} &@92.99 & --- &@86.34 & --- &@80.16 \\

  \bottomrule
    \end{tabular}
    \caption{UD 2.13 test sets LAS evaluation. Avg denotes the LAS macro average over the UD 2.13 corpora. Section E shows previous work on older UD versions.}
    \label{tab:ud2.13}
\end{table*}

\begin{table}
  \centering
  \catcode`! = 13\def!{\bfseries}
  \setlength{\tabcolsep}{2.15pt}  
  \renewcommand{\arraystretch}{0.945}
  \begin{tabular}{lccc}
      \toprule
      \textbf{Experiment} & Avg & \llap{P}oetr\rlap{y} & \llap{P}ros\rlap{e} \\
      \midrule
      
  \multicolumn{4}{l}{\sc A) Single-treebank Training}\vspace{0.1em}\\
    ~ITTB &  59.96 &  57.84 &  62.08 \\
    ~LLCT &  47.93 &  45.12 &  50.74 \\
    ~PROIEL original &  68.87 &  68.47 &  69.26 \\
    ~PROIEL harmonized & !73.88 & !72.37 & !75.40 \\
    ~UDante &  60.23 &  59.11 &  61.36 \\
    ~Perseus &  59.22 &  58.43 &  60.02 \\
  \midrule
  \multicolumn{4}{l}{\sc B) Multi-treebank with PROIEL Versions}\vspace{0.1em}\\
    ~UD 2.13, original &  72.31 &  72.10 &  72.52 \\
    ~UD 2.13, none &  66.16 &  64.03 &  68.29 \\
    ~UD 2.13, harmonized &  75.22 & !74.65 &  75.78 \\
    ~UD 2.13+Sab+Arch, original &  72.75 &  72.35 &  73.14 \\
    ~UD 2.13+Sab+Arch, none &  66.64 &  64.50 &  68.79 \\
    ~UD 2.13+Sab+Arch, harmo. & !75.48 &  74.52 & !76.43 \\
  \midrule
  \multicolumn{4}{l}{\sc C) Multi-treebank w/ and wo/ Gold UPOS}\vspace{0.1em}\\
    ~w/ gold UPOS & !75.48 & !74.52 & !76.43 \\
    ~wo/ gold UPOS &  74.19 &  73.28 &  75.09 \\
  \midrule
  \multicolumn{4}{l}{\sc D) Ensembles of 7 Models}\vspace{0.1em}\\
    ~UD 2.13+Sab+Arch, original &  73.76 &  73.57 &  73.95 \\
    ~UD 2.13+Sab+Arch, none &  68.16 &  65.71 &  70.60 \\
    ~UD 2.13+Sab+Arch, harmo. & !76.58 & !75.75 & !77.41 \\
  \midrule
  \multicolumn{4}{l}{\sc E) Adding Punctuation Before Prediction}\vspace{0.1em}\\
    ~UD 2.13+Sab+Arch, none &  71.87 &  70.68 &  73.07 \\
      \bottomrule
    \end{tabular}
    \caption{EvaLatin 2024 test set LAS evaluation. Avg denotes the LAS macro average over Poetry and Prose.}
    \label{tab:evalatin}
\end{table}

We present the evaluation on the UD 2.13 test data in Tables~\ref{tab:ud2.13} and on the EvaLatin 2024 test data in Table~\ref{tab:evalatin}. All the results are averages of three runs.

Table~\ref{tab:ud2.13}.A evaluates the baseline fine-tuned PLMs on the UD 2.13 test sets. Increasing PLM size from base to large clearly improves the results across the board and on average, even if the large model is not a monolingual but a multilingual one. The only exception is Perseus, on which we suspect the XLM-R large to overtrain due to the small size of the corpus (see Table~\ref{tab:training_data}). Finally, a concatenation of models yields further gains over their single components in all cases. 

Table~\ref{tab:ud2.13}.B shows a notable macro average gain of +1.42 percent points when pretraining with frozen weights for initial 10 epochs before fine-tuning. Also the addition of the two bidirectional LSTM layers helps marginally on average by +0.24. Unsurprisingly, the addition of gold UPOS on input brings +0.64 percent points in the UD 2.13 macro average, as well as it improves performance in all single UD 2.13 treebanks. On the EvaLatin test set, the addition of the gold UPOS straightforwardly improved the results by +1.2 on Poetry and +1.3 on Prose, as measured on the non-ensembled model (Table~\ref{tab:evalatin}.C). 

\begin{table*}
    \centering
    \catcode`! = 13\def!{\bfseries}
    \catcode`@ = 13\def@{\itshape}
    \def\0{\hphantom{0}}
    \setlength{\tabcolsep}{9.4pt}
    \renewcommand{\arraystretch}{0.945}
    \begin{tabular}{lcccccc}
      \toprule
      \textbf{Experiment} & Avg & ITTB & LLCT & \llap{P}ROIE\rlap{L} & \llap{U}Dant\rlap{e} & \llap{P}erseu\rlap{s} \\
      \midrule

  \multicolumn{7}{l}{\sc A) Best Single-model Results}\vspace{0.1em}\\
    ~~Single-treebank training & !97.33 & !99.37 & !99.77 & !98.32 & !93.61 &  95.55 \\
    ~~UD 2.13 training &  97.23 &  99.25 & !99.77 &  98.10 &  93.18 & !95.85 \\
  \midrule
  \multicolumn{7}{l}{\sc B) Best 7-Model Ensemble Results}\vspace{0.1em}\\
    ~~Single-treebank training, 7 models & !97.43 & !99.39 &  99.78 & !98.47 & !93.61 &  95.89 \\
    ~~UD 2.13 training, 7 models &  97.42 &  99.33 & !99.79 &  98.31 &  93.58 & !96.09 \\
    \midrule
    \multicolumn{7}{l}{\sc C) Previous work}\vspace{0.1em}\\
    \multicolumn{2}{l}{~~@UDPipe 2 \cite{straka-2018-udpipe}, UD 2.12} &@99.03 &@99.75 &@97.02 & @92.95 &@91.18 \\
    \multicolumn{2}{l}{~~@MaChAmp \cite{van-der-goot-etal-2021-massive}, UD 2.8} &@98.62 &@99.68 &@97.84 &@91.44 &@90.46 \\
    \multicolumn{2}{l}{~~@\citet{nehrdich-hellwig-2022-accurate}, UD 2.8-2.9} &@97.3\0 & --- &@94.2\0 & --- &@90.8\0 \\
    \multicolumn{2}{l}{~~@\citet{bamman2020latin}, UD 2.6} & @98.8\0 & --- &@98.2\0 & --- &@94.3\0 \\

  \bottomrule
    \end{tabular}
    \caption{UD 2.13 test sets UPOS evaluation, with Avg denoting the UPOS macro average.}
    \label{tab:ud2.13-upos-best}
\end{table*}

\begin{table*}
    \centering
    \catcode`! = 13\def!{\bfseries}
    \catcode`@ = 13\def@{\itshape}
    \def\0{\hphantom{0}}
    \setlength{\tabcolsep}{9.4pt}
    \renewcommand{\arraystretch}{0.945}
    \begin{tabular}{lcccccc}
      \toprule
      \textbf{Experiment} & Avg & ITTB & LLCT & \llap{P}ROIE\rlap{L} & \llap{U}Dant\rlap{e} & \llap{P}erseu\rlap{s} \\
      \midrule

  \multicolumn{7}{l}{\sc A) Best Single-model Results}\vspace{0.1em}\\
    ~~Single-treebank training &  92.45 & !98.57 &  97.33 & !94.68 &  83.06 &  88.61 \\
    ~~UD 2.13 training & !93.68 &  98.26 & !97.36 &  94.05 & !88.27 & !90.49 \\
  \midrule
  \multicolumn{7}{l}{\sc B) Best 7-Model Ensemble Results}\vspace{0.1em}\\
    ~~Single-treebank training, 7 models &  92.68 & !98.62 &  97.42 & !95.04 &  83.37 &  88.94 \\
    ~~UD 2.13 training, 7 models & !94.19 &  98.45 & !97.52 &  94.56 & !89.16 & !91.24 \\
    \midrule
    \multicolumn{7}{l}{\sc C) Previous work}\vspace{0.1em}\\
    \multicolumn{2}{l}{~~@UDPipe 2 \cite{straka-2018-udpipe}, UD 2.12} &@97.12 &@97.16 &@91.43 & @84.38 &@84.65 \\
    \multicolumn{2}{l}{~~@MaChAmp \cite{van-der-goot-etal-2021-massive}, UD 2.8} &@96.95 &@96.79 &@92.56 &@69.72 &@84.32 \\

  \bottomrule
    \end{tabular}
    \caption{UD 2.13 test sets UFeats evaluation, with Avg denoting the UFeats macro average.}
    \label{tab:ud2.13-ufeats-best}
\end{table*}

Table~\ref{tab:ud2.13}.C compares multi-treebank training vs. single-treebank training. In accord with previous literature \cite{nehrdich-hellwig-2022-accurate}, we observed the greatest benefits from the multi-treebank training for the smaller datasets (UDante and Perseus), indecisive results for the middle-sized datasets (LLCT and PROIEL), and a decrease for the largest dataset (ITTB). However, in macro average, we gained +0.51 percent point by multi-treebank training. While the addition of the two new small datasets, the Sab and Arch, is indecisive on the UD 2.13 macro average in Table~\ref{tab:ud2.13}.C, which is in alignment with their modest size (Table~\ref{tab:training_data}), on EvaLatin 2024 (Table~\ref{tab:evalatin}.B), we observed a marginal improvement when incorporating Sab and Arch, which might probably be attributed to similarity of the EvaLatin test data to these treebanks.

Table~\ref{tab:evalatin} shows the evaluation on the EvaLatin test data, both Poetry and Prose, and their LAS macro average; with focus on the effect of data harmonization. In all paired experiments, the harmonized PROIEL version clearly improved results over the version with the original PROIEL dataset from UD 2.13, when evaluated on the EvaLatin 2024 test data. However, using at least the original PROIEL dataset in the multi-treebank training is still better than excluding the PROIEL treebank altogether.

As evidenced by both Table~\ref{tab:ud2.13}.D and Table~\ref{tab:evalatin}.D, an ensemble is always stronger than its individual components. Ensembling adds on average +0.45 percent points on the UD 2.13 LAS macro average over three experimental settings (compare sections C and D in Table~\ref{tab:ud2.13}). In the shared task, ensembling adds +1.26 percent points (compare sections B and D in Table~\ref{tab:evalatin}). Our best entry, submitted as \textit{ÚFAL LatinPipe~1}, corresponds to the row \textit{UD 2.13+Sab+Arch, harmo.} in Table~\ref{tab:evalatin}.D.

Finally, when training without PROIEL in a multi-treebank setting, we have to mitigate the punctuation mismatch between the training and the shared task test data, as described in Section~\ref{sec:methods}. Row \textit{UD 2.13+Sab+Arch} in Table~\ref{tab:evalatin}.E shows our second submission to the shared task, \textit{ÚFAL LatinPipe2}, in which we corrected for missing punctuation in the shared task test data.

\paragraph{UPOS and UFeats Tagging} Since our model performs full morphosyntactic analysis, we present also the accuracy of UPOS tagging and UFeats tagging in Tables~\ref{tab:ud2.13-upos-best}~and~\ref{tab:ud2.13-ufeats-best}, respectively. \latinpipe surpasses the previous systems and sets new state-of-the-art results for all treebanks.

\section{Conclusion}
\label{sec:conclusion}

We described \latinpipe, the winning entry to the \evalatin, and we provided the evaluation and rationale behind our system design choices. The source code for \latinpipe is available at {\small\url{https://github.com/ufal/evalatin2024-latinpipe}}. Our future work will entail drawing insights from the methodologies presented in this context for the development of UDPipe 3.

\section{Acknowledgements}

This work has been supported by the Grant Agency of the Czech Republic under the EXPRO program as project “LUSyD” (project No.\ GX20-16819X), and by the Grant Agency of Charles University as project GAUK No.\ 104924 ``Adapting Uniform Meaning Representation (UMR) for the Italic/Romance languages''. The work described herein uses resources hosted by the LINDAT/CLARIAH-CZ Research Infrastructure (projects LM2018101 and LM2023062, supported by the Ministry of Education, Youth and Sports of the Czech Republic). 

\nocite{*}
\section{Bibliographical References}
\label{sec:reference}
\bibliographystyle{lrec-coling2024-natbib}
\bibliography{lrec-coling2024}

\section{Language Resource References}
\label{sec:languageresource}
\bibliographystylelanguageresource{lrec-coling2024-natbib}
\bibliographylanguageresource{languageresource}

\end{document}